\documentclass{article}
\usepackage{ijcai24}

\usepackage{times}
\usepackage{soul}
\usepackage{url}
\usepackage[colorlinks]{hyperref}
\usepackage[utf8]{inputenc}
\usepackage[small]{caption}
\usepackage{graphicx}
\usepackage{amsmath}
\usepackage{amsthm}
\usepackage{amssymb}
\usepackage{booktabs}
\usepackage{algorithm}
\usepackage{algorithmic}
\usepackage[switch]{lineno}
\usepackage{bm}
\usepackage{xcolor}
\usepackage{tabularx}
\usepackage{multirow}

\usepackage{pifont} 
\usepackage{amssymb} 
\usepackage{pifont}
\usepackage{ragged2e}

\title{Federated Learning with New Knowledge: Fundamentals, Advances, and Futures}

\author{
\normalsize
Lixu Wang$^{1}$\thanks{These authors contributed equally to this work.}
\and
Yang Zhao$^{2,6}$\footnotemark[1]\and
Jiahua Dong$^{3}$\thanks{Corresponding to Jiahua Dong.}
\and
Ating Yin$^4$\and
Qinbin Li$^5$\and
Xiao Wang$^1$\and
Dusit Niyato$^6$ \And
Qi Zhu$^1$
\\
\affiliations
\normalsize
$^1$Northwestern University, USA. \quad
$^2$Agency for Science, Technology and Research (A$*$STAR), Singapore.\\
$^3$Shenyang Institute of Automation, Chinese Academy of Sciences, China. \quad $^4$Hunan University, China.\\
$^5$University of California, Berkeley, USA. \quad
$^6$Nanyang Technological University, Singapore. 
}

\begin{document}

\maketitle

\begin{abstract}
Federated Learning (FL) is a privacy-preserving distributed learning approach that is rapidly developing in an era where privacy protection is increasingly valued. It is this rapid development trend, along with the continuous emergence of new demands for FL in the real world, that prompts us to focus on a very important problem: \emph{Federated Learning with New Knowledge}. The primary challenge here is to effectively incorporate various new knowledge into existing FL systems and evolve these systems to reduce costs, extend their lifespan, and facilitate sustainable development. 
In this paper, we systematically define the main sources of new knowledge in FL, including new features, tasks, models, and algorithms. For each source, we thoroughly analyze and discuss how to incorporate new knowledge into existing FL systems and examine the impact of the form and timing of new knowledge arrival on the incorporation process. Furthermore, we comprehensively discuss the potential future directions for FL with new knowledge, considering a variety of factors such as scenario setups, efficiency, and security. There is also a continuously updating repository for this topic: \href{https://github.com/conditionWang/FLNK}{https://github.com/conditionWang/FLNK.}
\end{abstract}


\section{Introduction}



Federated Learning (FL), as an emerging form of distributed learning, has received significant attention and achieved breakthrough developments over the past decade. This surge in interest is primarily due to its ability to facilitate multi-party collaborative training of machine learning (ML) models while protecting the privacy of training data. It is exactly because of such characteristics of privacy protection that an increasing number of critical areas, such as medical healthcare, intelligent transportation, financial investment, digital commerce, and social media, have embraced FL and introduced a series of real-world applications~\cite{fl_survey,fl_survey_2,ratio}.

The aforementioned practical applications reflect a tremendous demand for FL, spurring its rapid development and evolution. In the field of ML, technological advancements often result in the withdrawal of old models and the introduction of new ones. However, given a large number of participating clients, both in terms of data volume, computation overhead, and communication bandwidth, training of each FL model actually incurs substantial costs~\cite{fl_survey_2}. Furthermore, FL applications are often situated in rapidly changing environments, such as the emergence of new diseases, autonomous vehicles encountering unfamiliar roads, or the appearance of new investment trends~\cite{fedweit,glfc,Peng2019FederatedAD,li2023federated,yu2023turning}. In addition to the introduction of these new concepts, there may be sensor degradation, hardware damage, product evolution, and so on, in FL deployment scenarios. However, current FL systems typically assume a fixed and predetermined distribution of data and tasks~\cite{fl_survey,fl_survey_2}, making it challenging to deal with dynamic changes in data and tasks in real-world scenarios. These practical considerations reflect a dilemma in the future development of FL: on the one hand, there is a desire to establish new FL systems in response to emerging demands and technologies, but on the other hand, the significant investment in established FL systems makes it impractical to discard them readily. Therefore, how to achieve sustainable evolution in FL, and extending the lifespan of FL systems becomes an extremely important research direction.

Fortunately, there are more and more studies starting to explore relevant problems, mainly intending to propose new technical solutions to help FL learn new changes in application scenarios. For instance, \cite{sun2023feature,Peng2019FederatedAD} achieve domain-generalizable learning to improve FL models' generalization ability, and \cite{glfc,qi2022better} incorporate continual learning techniques to learn new tasks. However, there is a lack of a survey that reviews these potential works from the perspective of expanding the lifespan of FL, which can inspire new thinking about FL's environmental sustainability, continual functionality iteration, and temporal cost reduction.

\begin{figure*}[htbp]
    \centering
    \includegraphics[width=.99\textwidth]{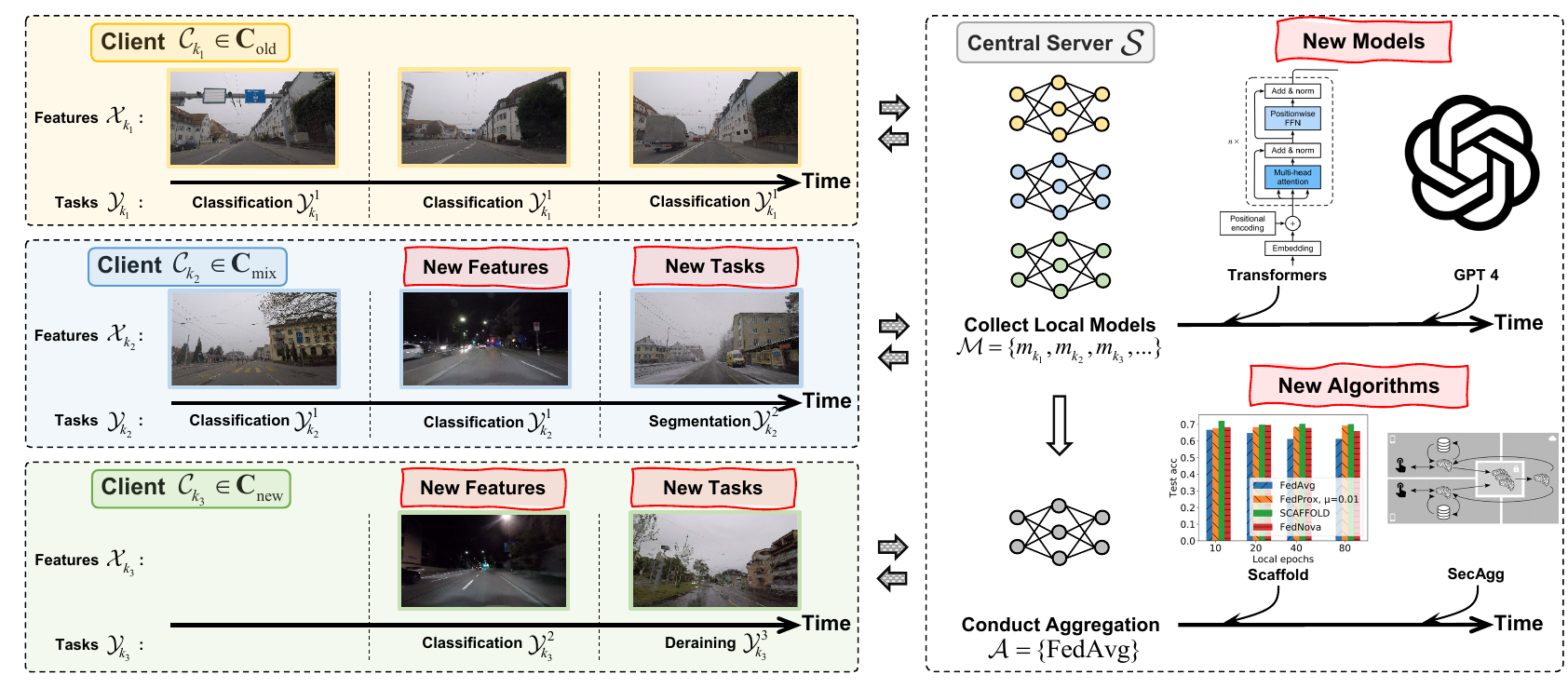}
    \vspace{-4mm}
    \caption{Overview of an FL system with new knowledge from different sources. Different types of clients encounter new features and tasks over time, which reflect new demands for FL systems, e.g., client $\mathcal{C}_{k_2}$ needs to deal with the night scenes and conduct segmentation when snowing, and client $\mathcal{C}_{k_3}$ joins FL with the need of handling night scenes and deraining when raining. From a global view, new more advanced models with better architecture (Transformers) and larger sizes (GPT 4) are also needed to incorporate. Besides, new algorithms with better performance (Scaffold) and security guarantees (SecAgg) should be continuously employed in FL as well.}
    \label{fig_overview}
\end{figure*}

\smallskip \noindent \textbf{Contributions:}
In this paper, we present the first systematic and comprehensive survey with analysis and discussion on how FL can achieve continuous updates and development in the face of new knowledge.  More importantly, we define the FL system across four variables: features, tasks, models, and algorithms. For each variable, we provide a detailed analysis and discussion on what its corresponding new knowledge looks like and how to incorporate such knowledge into the current FL system in a timely and efficient manner. We also analyze an impact of the arrival form and timing of the new knowledge on its incorporation into the FL system. Moreover, we extensively discuss future research priorities for the continuous development of FL in particular with the incorporation of various new knowledge.

\begin{table*}[!h]
\centering
\resizebox{1\textwidth}{!}{%
\small
\setlength{\tabcolsep}{.5mm}{
\begin{tabularx}{\textwidth}{lllccc}
\toprule
\multicolumn{1}{c}{\textbf{Knowledge Type}}& \multicolumn{1}{c}{\textbf{Method}}&\multicolumn{1}{c}{\textbf{State-of-the-art}} &\multicolumn{1}{c}{\textbf{Metric}}  & \multicolumn{1}{c}{\textbf{Dataset}} & \multicolumn{1}{c}{\textbf{Results}} \\
\hline
\multirow{3}{*}{\ref{sec_new_features}~New Features}  & \ref{sec_fdg}~Federated Domain Generalization  &\cite{FDG_sota}  & Leave-1-out Acc & PACS & 86.2\%\\
  & \ref{sec_fodd}~Federated Out-of-Distribution Detection& \cite{yu2023turning} & AUROC & CIFAR-10 & 0.909 \\
    & \ref{sec_fda}~Federated Domain Adaptation& \cite{fda_sota} & Leave-1-out Acc & DomainNet & 51.1\% \\
\hline
\multirow{2}{*}{\ref{sec_new_tasks}~New Tasks}  &\ref{sec_task_personalized_fl}~Task-Personalized Federated Learning &\cite{huang2021personalized}  & Test Acc & FMNIST & 91.37\%\\
  &\ref{sec_ssfl} Self-Supervised Federated Learning& \cite{zhuang2022divergence} & Top-1 Acc &  CIFAR-10 & 86.1\%  \\
\hline
\multirow{3}{*}{\parbox{3cm}{\ref{sec_new_tasks_new_features} New Tasks with\\New Features}}  & \ref{sec_fcl}~Federated Continual Learning&\cite{glfc}  & ACC, FGT & ImageNet & 57.0, 7.8\%\\
  & & \cite{anonymous2023accurate} & ACC, FGT & CIFAR-100 & 36.3, 4.9\% \\
  & & \cite{fot} & ACC, FGT & ImageNet & 62.1, 0.2\% \\
\hline
\multirow{1}{*}{\ref{sec_future_direction}~Future Directions}&Potential Research Focuses&&\\
\hline
\ref{sec_future_features_tasks}~New Features and Tasks& \multicolumn{5}{l}{\ding{172}Multi-Label and
Multi-Grained Classification \ding{173}More Complicated Tasks \ding{174}Efficiency Improvement~$\cdots$}\\
\ref{sec_new_models}~New Models& \multicolumn{5}{l}{\ding{172}Incorporate Foundation or Large Language Models into FL \ding{173}Transfer Existing FL Models to New Models}\\ 
\ref{sec_new_algorithms}~New Algorithms& \multicolumn{5}{l}{\ding{172}Speed Up Convergence \ding{173}Optimize Timing of Switching Algorithms \ding{174}Reduce Communication Cost~$\cdots$} \\
\ref{sec_new_threats}~New Threats & \multicolumn{5}{l}{\ding{172}Privacy Protection in Semi-Honest Cases \ding{173}Poison Attacks \ding{174}Backdoor Attacks \ding{175}New Attack Surfaces~$\cdots$} \\
  \hline
\end{tabularx}}}
\vspace{-2mm}
\caption{Organization Overview and State-of-the-Arts Preview of Federated Learning with New Knowledge.}
\label{Table_Roadmap}
\end{table*}






\section{Fundamentals and Taxonomy}
Federated Learning (FL)~\cite{fl_survey}, as a distributed learning algorithm, enables multiple clients $\{\mathcal{C}_k\}_{k=1}^K$ to collaboratively train a machine learning model $m:f_\theta \!\circ\! g_\omega$ without sharing their training data $\mathcal{D}_k \!=\! \{(\bm{x}_{k, i}, y_{k, i}) \sim (\mathcal{X}_k, \mathcal{Y}_k)\}_{i=1}^{N_k}$. In FL, clients possess their training data in a non-independent and identically distributed (non-IID) manner, i.e., the feature and task marginal distributions of different clients are not identical, $\mathcal{X}_k \neq \mathcal{X}_{k^\prime}, \mathcal{Y}_k \neq \mathcal{Y}_{k^\prime}$ where $k \neq k^\prime$. A typical federated training round begins with a central server $\mathcal{S}$ distributes the latest global model $m$ to each client. Additionally, the server randomly selects a subset of clients $\{\mathcal{C}_k\}_{k=k_1}^{k_s}$ and requests them to conduct model training locally using their own data. Once local training is completed, these selected clients upload their locally trained models $m_k^\prime$ to the server $\mathcal{S}$. The server then aggregates these models $\{m_k^\prime\}_{k=k_1}^{k_s}$ to update the global model, and the most popular aggregation algorithm is FedAvg~\cite{fedavg}, i.e., $m^\prime = \frac{1}{s}\sum_{k=k_1}^{k_s}m_k^\prime$. This process is repeated over multiple rounds until the global model converges and exhibits satisfactory performance for a given task. 

From the description above, we can define different FL systems $F(\mathcal{X}, \mathcal{Y}, \mathcal{M}, \mathcal{A})$ using the following variables: features $\mathcal{X} \!=\! \cup_{k=1}^K\mathcal{X}_k$, tasks $\mathcal{Y} \!=\! \cup_{k=1}^K\mathcal{Y}_k$, models $\mathcal{M}\!=\!\{(f_\theta, f_\alpha, \cdots) \circ g_\omega, f_\theta \circ (g_\omega, g_\beta, \cdots)\}$, and aggregation algorithms $\mathcal{A}\!=\!\{$FedAvg, FedProx~\cite{fedprox}, Moon~\cite{moon}, SecAgg~\cite{secagg}, CaPC~\cite{capc}$, \cdots\}$.

\smallskip \noindent \textbf{Lifespan of Federated Learning.} Suppose there are $K$ clients in total at the beginning of an FL system $F(\mathcal{X}, \mathcal{Y}, \mathcal{M}, \mathcal{A})$, the lifespan $T$ of this system is defined as the period from the start until when the client number is decreased to $0.1K$. Note that there will be dynamic participation and withdrawal of clients, including those who only conduct FL model inference. The lifespan $T$ includes both the training and inference periods, and we assume clients are free to join or withdraw at any time, for example, if an FL system spends too much time on training, clients may choose to withdraw as their demands cannot be satisfied in time.

\smallskip \noindent \textbf{Federated Learning with New Knowledge.} Suppose the lifespan of an FL system $F(\mathcal{X}, \mathcal{Y}, \mathcal{M}, \mathcal{A})$ be denoted as a sequence of time stamps $[1, 2, ..., t, ..., T]$, then the new knowledge is defined as a particular variable at a particular timestamp that is unseen previously. If we take the feature variable as an example, we have $\mathcal{X}^t \nsubseteq \cup_{i=1}^{t-1}\mathcal{X}^i$. Note that the arrival of new knowledge in FL can occur during training or post-training, and such new knowledge is often brought by different clients. To facilitate the coming of more new knowledge, we do not restrict clients' participation, which means clients are free to join or leave the FL system at any time. Under this framework, we categorize the FL clients into three categories: $\mathbf{C}_\mathrm{old}\!=\!\{\mathcal{C}_k\}_{k=k_{\mathrm{old}, 1}}^{k_{\mathrm{old}, n_\mathrm{old}}}$ corresponds to clients that possess only old knowledge; $\mathbf{C}_\mathrm{new}\!=\!\{\mathcal{C}_k\}_{k=k_{\mathrm{new}, 1}}^{k_{\mathrm{new}, n_\mathrm{new}}}$ corresponds to clients that only possess new knowledge; $\mathbf{C}_\mathrm{mix}\!=\!\{\mathcal{C}_k\}_{k=k_{\mathrm{mix}, 1}}^{k_{\mathrm{mix}, n_\mathrm{mix}}}$ corresponds to clients that possess both old and new knowledge. Note that the category belonging of a particular client $\mathcal{C}_k$ is changing over time, thus these three client groups $\mathbf{C}_\mathrm{old}, \mathbf{C}_\mathrm{new}$, and $\mathbf{C}_\mathrm{mix}$ are also changing dynamically. Overall, regardless of which variable the new knowledge belongs to, when it arrives, or who introduces it, we need to incorporate it promptly. Incorporating such new knowledge is crucial because the arrival of new knowledge essentially signifies clients' new demands for enhancing the performance and functionality of the current FL system. We visualize an FL system with various kinds of new knowledge in Figure~\ref{fig_overview}. 

\smallskip \noindent \textbf{Threat Model of FL with New Knowledge.}
A trusted setting is assumed when incorporating new knowledge into an FL system, i.e., 1) all participants including both the central server $\mathcal{S}$ and clients $\{\mathcal{C}_k\}_{k=1}^K$ are trusted to not eavesdrop private information or launch active attacks in both semi-honest and malicious scenarios; 2) all carriers of new knowledge including features $\mathcal{X}$, tasks $\mathcal{Y}$, models $\mathcal{M}$, and algorithms $\mathcal{A}$ don't intend to disclose private information or contain active attacks in both semi-honest and malicious scenarios. 

The reason why we consider such trusted settings is that we aim to discuss the technical advances of FL with new knowledge in the main context and we found all existing related works assume such trusted scenarios. We also talk about new threats in both semi-honest and malicious settings, please refer to Section~\ref{sec_new_threats} for more details. In the forthcoming sections, we organize current research on Federated Learning with New Knowledge following the four variables mentioned earlier. The detailed organization and a high-level preview of the advances we review can be found in Table~\ref{Table_Roadmap}. 

\section{Incorporate New Features}
\label{sec_new_features}



One of the most representative applications of FL is to develop ML models across edge devices like mobile phones, IoT, and wearables. When such models are deployed on new devices that usually work on unseen new features, the model performance degrades significantly due to the large domain discrepancy between the learned features (source domain) and unseen new features (target domain). Therefore, it is important to improve the generalization ability of FL systems on unseen new features. Specifically, we consider that Federated Domain Generalization (FDG) can equip the FL system with a certain degree of generalization ability on new unseen features. Federated Out-of-Distribution Detection (FODD) can detect the arrival of new features, and then determine the need of incorporating new features. After that, Federated Domain Adaptation (FDA) conducts adaptive learning on unlabeled data with new features to minimize domain discrepancy, achieving the incorporation of new features.



\subsection{Federated Domain Generalization}
\label{sec_fdg}

FDG aims to train a domain-generalizable global model $m$ that can generalize well across different distributions. Such domain-generalizable learning is usually achieved by extracting domain-shared semantics across local clients with old knowledge $\mathbf{C}_\mathrm{old}, \mathbf{C}_\mathrm{mix}$ (source domains). In this way, model $m$ can extract similar semantics from the unseen new features $\mathcal{X}_\mathrm{new}$. Here we assume that $\mathcal{X}_\mathrm{new}$ is relevant but distinct to $\cup_{k=1}^K \mathcal{X}_k$ in terms of the same task $\mathcal{Y}$.


\smallskip \noindent \textbf{Methods -- Distribution Alignment.} Several FDG methods~\cite{li2023federated} employ feature alignment techniques to minimize domain discrepancy, thus improving model generalization across diverse domains.
Generally, they achieve this improvement by optimizing with various distribution alignment regularizers~\cite{xu2021weak}, such as contrastive loss, adversarial loss, and Maximum Mean Discrepancy (MMD) loss.
For example, \cite{xu2023federated} propose a negative-free contrastive loss at the logits level to minimize the distribution gap between the original sample and its hallucinated counterpart.
Regarding the adversarial loss, \cite{wang2022federated} employs adversarial learning to perform multi-client feature alignment to acquire domain generalization features from diverse clients.
Similarly, \cite{zhang2023federated} introduces another generalization adjustment model via dynamically calibrating aggregation weights with an adversarial objective. 
As for the MMD aspect, \cite{tian2023privacy} utilizes MMD to perform gradient alignment, which encourages the aggregated gradients to unify information from multiple domains.
In addition, \cite{sun2023feature} introduces federated knowledge alignment to learn domain-invariant client features while mitigating negative transfer within FL.

\smallskip \noindent \textbf{Data Augmentation.}
To enhance the model's generalization ability to unseen target domains, other FDG methods leverage data augmentation techniques to diversify the training data distributions. In this way, if the unseen data domains are similar to the augmented training data, the model generalizes better on them~\cite{liu2022deja}. Generally, such augmentation techniques include domain randomization, domain generation, and domain mixup~\cite{liu2022deja}. Domain randomization is used to diversify the data variability across different clients via randomizing specific attributes. For example, \cite{liu2021feddg} utilizes domain randomization techniques including random rotation, scaling, and flipping to solve the FDG problem.
As for domain translation, \cite{chen2023federated} introduces a cross-client style transfer method, where local clients evaluate their local styles and share them with a central server. 
Besides, to generate diverse domains via domain mix-up technology, \cite{yang2023clientagnostic} proposes to mix local and global feature statistics by randomly interpolating instance and global statistics.

\subsection{Federated Out-of-Distribution Detection}
\label{sec_fodd}

FODD was proposed to detect the appearance of new features across local clients (usually $\mathbf{C}_\mathrm{mix}$ and $\mathbf{C}_\mathrm{new}$) using a binary classifier $g_b$ that is attached behind the global feature extractor $f_\theta$. Once the data with new features is detected, there is supposed to be a need to incorporate them into the current FL system. Therefore, $g_b$ should be trained to tell us whether a sample $\bm{x}$ is drawn from previously unseen feature distribution $\bm{x} \sim \cup_{i=1}^{t-1} \mathcal{X}_i$. Relevant studies are reviewed below.


\smallskip \noindent \textbf{Methods.}
\cite{nardi2022anomaly} views the FODD problem as achieving anomaly detection in a decentralized manner, in which all clients are initially clustered into communities based on their extracted representation similarity, and then they collaboratively train a federated autoencoder to detect anomalies.
Similarly, \cite{Mohammadi2023} employs a modified autoencoder with a long short-term memory mechanism involved to detect outlier data in FL. \cite{glfc,dong2023federated_FISS} proposes a task transition monitor module to detect out-of-distribution learning tasks.
\cite{yu2023turning} trains a generator specific to each data category to synthesize out-of-distribution samples for training stronger FODD models. \cite{fedcn} proposes a hierarchical clustering mechanism to discover and learn novel features that haven't been learned previously.

\subsection{Federated Domain Adaptation}
\label{sec_fda}
FDA aims to collaboratively adapt the global model on the data with new features detected by FODD. In FDA, we assume there is a large domain discrepancy between the old and new features, $\mathrm{d}(\mathcal{X}_\mathrm{new}, \mathcal{X}_k) \gg \max{d(\mathcal{X}_k, \mathcal{X}_{k^\prime}})$ ($d(\cdot)$ denotes a function to measure distribution discrepancy), thus the major problem is still how to minimize such discrepancy. As most data with new features appears with no label, FDA should address domain discrepancy \cite{9616392_Dong,What_Transferred_Dong_CVPR2020} on both labeled source data with old features and unlabeled target data with new features. To achieve that, current FDA studies usually leverage adversarial training and pseudo-label learning.


\smallskip \noindent \textbf{Methods -- Adversarial Training.}
\cite{Peng2019FederatedAD} first builds upon adversarial adaptation techniques tailored to the FDA's constraints and incorporate a dynamic attention mechanism to bolster the effectiveness of knowledge transfer. Then \cite{9706703} considers tackling both server-client and inter-client domain shifts, and defines a new challenge named Federated Multi-Target Domain Adaptation that involves a single source and multiple targets. Similarly, \cite{9965364} proposes two stages to minimize domain discrepancy: a pre-training stage to perform adversarial adaptation with gradient matching loss and a fine-tuning stage to update local models. Furthermore, \cite{9376674} proposes a federated transfer learning framework to tackle fault diagnosis via deep adversarial learning.

\smallskip \noindent \textbf{Pseudo-label Learning.} To generate confident pseudo labels for unlabeled target data with new knowledge, \cite{9710554} introduces a collaborative optimization and weighted aggregation strategy to produce pseudo labels by comparing the distance to global prototypes. \cite{9859587} aims to tackle face recognition tasks by integrating clustering-based pseudo labels into FDA. Moreover, \cite{ZHAO2023109246} applies the FDA to perform machinery fault diagnosis with data privacy, and the authors focus on alleviating the negative transfer during feature alignment and then generate pseudo-labeled samples to refine fault diagnosis results.



\section{Incorporate New Tasks}
\label{sec_new_tasks}


In the dynamic field of FL, conventional systems encounter significant challenges when addressing the diverse and continuously emerging tasks presented by clients. The introduction of new functionalities, which can be comparable to new tasks with similar features, is a frequent phenomenon. To effectively respond to these evolving requirements, it is critical to enhance the cross-task generalization capabilities of FL models. This necessity directs our attention towards advanced methodologies in FL, with a specific emphasis on task-personalized FL and self-supervised FL. 

\subsection{Task-Personalized Federated Learning}
\label{sec_task_personalized_fl}
Task-Personalized FL is essential for addressing the diverse and specific requirements of clients within a federated network. This approach deals with the inherent diversity in feature and task distributions among different clients. Each client  $\mathcal{C}_k$ adapts the global model $m:f_\theta \circ g_\omega$ to better align with its local data characteristics and specific needs. Initially, the central server $\mathcal{S}$ distributes the global model to each client. The clients then use their unique datasets $\mathcal{D}_k$ to locally train models $m_k^\prime$, optimizing a personalized objective function 
$\mathcal{L}_k\left[m_k^\prime(\mathcal{X}_k), \mathcal{Y}_k)\right]$. 
These models are aggregated by the server using FL algorithms from set $\mathcal{A}$, resulting in an updated global model $m^\prime$ with better cross-task generalization ability due to this multi-task training process. 

\smallskip \noindent \textbf{Methods.}
Some preliminary studies include FedAM~\cite{huang2021personalized} and Ditto~\cite{li2021ditto},
which allow for maintenance and updates of personalized models in FL. These methods enable clients to tailor their models based on information from other clients or through efficient, privacy-preserving personalization techniques. Similarly, the pFedMe framework~\cite{t2020personalized} introduces a novel approach in FL by framing a bi-level optimization problem for personalized FL, using regularized loss functions like the Moreau envelop, which particularly addresses client-specific errors and ensures convergence in personalized settings. Moreover, methods like MOCHA, proposed by~\cite{smith2017federated}, and VIRTUAL, by~\cite{corinzia2019variational}, focus on individual model learning while concurrently addressing broader statistical challenges in FL. This aspect is crucial for enhancing personalization as it considers each client's unique contribution to the federated network. Furthermore, techniques that leverage frameworks like MAML, as highlighted in work~\cite{fallah2020personalized}, are designed to create adaptable shared initial models for FL clients.

\subsection{Self-Supervised Federated Learning}
\label{sec_ssfl}

Self-Supervised FL (SSFL) stands out as another crucial approach in improving the cross-task generalization ability of FL models, especially in environments where labeled data is scarce or unavailable. The key difference from the task-personalized FL lies in how each client $\mathcal{C}_k$ utilizes its local data $\mathcal{D}_k$. More importantly,
here each client independently creates its own self-supervised learning tasks, such as contrastive learning, rotation prediction, and so on. Specifically,
client $\mathcal{C}_k$ then uses a unique objective function: $\mathcal{L}_{\mathrm{self}}\left[m_k^\prime(\mathcal{X}_k)\right]$ to optimize its local model $m_k^\prime$, where $\mathcal{L}_{\mathrm{self}}$ is a self-supervised learning (SSL) loss function, such as infoNCE loss. Overall, SSFL enables the extraction and integration of diverse, self-generated features and knowledge across clients, which is beneficial for incorporating new tasks.


\smallskip \noindent \textbf{Methods.}
A preliminary SSFL framework proposed by~\cite{he2021ssfl} integrates SSL strategies to facilitate both collaborative and personalized model training. Additionally, SS-VFL by~\cite{castiglia2022self} extends this approach to Vertical FL, leveraging unlabeled data for enhancing representation networks alongside labeled data for prediction tasks. Moreover, MocoSFL, proposed by~\cite{li2022mocosfl}, combines split FL and momentum contrast in an SSL framework, targeting computational and data challenges in non-IID environments. This approach has demonstrated effectiveness in achieving high accuracy and communication efficiency.~\cite{liang2021self} further validates the effectiveness of SSFL on both synthetic and real non-IID datasets, showing that representation regularization-based personalization methods excel in these settings. Finally, FedEMA, introduced by~\cite{zhuang2022divergence}, presents a generalized framework for non-IID decentralized data, focusing on preserving local client knowledge crucial for adaptability to new tasks. 

\section{Incorporate New Tasks with New Features}

\label{sec_new_tasks_new_features}
In practical applications of FL, the arrival of new tasks is often accompanied by new features. Typically, learning new tasks based on the models that have already undergone some training is referred to as continual learning (CL). As a result, it is necessary to extend CL to the FL context, achieving Federated Continual Learning (FCL)~\cite{yang2023federated}. While regular CL is not limited to classification, existing research in FCL nearly all assumes that new tasks are presented as new data classes. Therefore, we use classification as a representative example to elucidate FCL here.

\subsection{Federated Continual Learning}
\label{sec_fcl}
\noindent \textbf{Synchronous FCL.} Similar to standard CL, Synchronous FCL assumes that there are a series of datasets $\{\mathcal{D}^t\}_{t=1}^T$ with $N^t$ data samples $\mathcal{D}^t = \{(\bm{x}_j^t, y_j^t)\}_{j=1}^{N^t}$ continuously arriving in the FL systems globally. The majority of each dataset is assumed to belong to previously unseen data classes, i.e., $\mathcal{Y}^t \cap \cup_{i=1}^{t-1}\mathcal{Y}^i = \emptyset$. Moreover, actually, the dataset at each time $t$ is also non-IID owned by a part of clients at that time, i.e., each client $\mathcal{C}_k$ possesses $\mathcal{D}_k^t = \{(\bm{x}_{k, j}^t, y_{k, j}^t)\}_{j=1}^{N^t_k}$ and the corresponding marginal distributions are distinct among clients, $\mathcal{X}_k^t \neq \mathcal{X}_{k^\prime}^t, \mathcal{Y}_k^t \neq \mathcal{Y}_{k^\prime}^t$ (where $k \neq k^\prime$, but $\mathcal{Y}_k^t, \mathcal{Y}_{k^\prime}^t \subseteq \mathcal{Y}^t$). Consistent with existing Synchronous FCL studies, we also consider practical constraints from limited storage space and privacy regulations~\cite{ntl,guo2023domain}, and assume that old class data is unavailable or can be only accessed partially by the clients that own new class data ($\mathbf{C}_\mathrm{new}$ and $\mathbf{C}_\mathrm{mix}$). Then the objective is to minimize the classification errors for new classes while preserving the good performance of learned classes
\begin{equation}
\begin{aligned}
\min\limits_{\theta^t, \omega^t}\, &\mathbb{E}_{\bm{x} \sim \cup_{i=1}^{t-1}\mathcal{X}^i}\left[\|g_{\omega^t}(f_{\theta^t}(\bm{x})) \!-\! g_{\omega^{t-1}}(f_{\theta^{t-1}}(\bm{x}))\|^2\right] \\
+\,&\mathbb{E}_{(\bm{x}, y) \sim (\mathcal{X}^t, \mathcal{Y}^t)} \left[\mathcal{L}(g_{\omega^t}(f_{\theta^t}(\bm{x})), y)\right].
\label{eq_synchronous_FCL}
\end{aligned}
\end{equation}

\smallskip \noindent \textbf{Asynchronous FCL.} Different from Synchronous FCL, Asynchronous FCL assumes that FL clients learn a series of classes in their own distinct orders. Specifically, suppose there is a set of class spaces $\{\mathcal{Y}^t\}_{t=1}^T$ with the order $\mathcal{O} = [1, ..., t, ..., T]$ from the global perspective, and these class spaces are disjoint from each other $\mathcal{Y}^t \cap \mathcal{Y}^{t^\prime} = \emptyset$ (where $t \neq t^\prime$). However, from the local perspective, each client $\mathcal{C}_k$ continually receives the dataset corresponding to a particular order $\mathcal{O}_k = [1_k, ..., t_k, ..., T_k]$. The orders of clients are distinct from each other as well as from the global one, i.e., $\mathcal{O}_k \neq \mathcal{O}_{k^\prime}, \mathcal{O}_k \neq \mathcal{O}$ (where $k \neq k^\prime$). Asynchronous FCL also conforms to the constraints of limited storage space and privacy regulations. As for the difference between different client categories, we assume $\mathbf{C}_\mathrm{old}$'s data is fixed until the category changes into others, while the data of both $\mathbf{C}_\mathrm{new}$ and $\mathbf{C}_\mathrm{mix}$ is changing over time as long as they do not change into $\mathbf{C}_\mathrm{old}$. Then the objective of Asynchronous FCL is to minimize the classification errors for new classes without or with partial access to previously learned classes on the local side, which can be viewed as a local personalization of Eq.~(\ref{eq_synchronous_FCL}).

\subsection{Methods}
\noindent \textbf{Regularization-based Approaches.} During CL, the model parameters are optimized for new tasks, resulting in the forgetting of learned tasks. Adding additional regularization to prevent large parameter changes is a straightforward way of alleviating forgetting. Fisher information and synaptic intelligence are commonly used to measure the importance of model parameters, and then FedCurv~\cite{fedcurv} and CPPFL~\cite{cppfl} are pioneers in applying them to the local training of FL. However, these two works do not consider the impact of data heterogeneity in FL, which is the main focus of \cite{ewc_extension}. 
Model parameter change is also reflected as intermediate layer output change, thus another forgetting alleviation is to conduct knowledge distillation for preserving learned task knowledge. FLwF~\cite{flwf} first builds a teacher-student framework among a stored old task model and the current model, and regularizes similar logits over the same input, which corresponds to the first term of Eq.~(\ref{eq_synchronous_FCL}). GLFC~\cite{glfc} designs a weighting mechanism for logit regularization to alleviate the impact of data heterogeneity. 
By contrast, FCCL~\cite{fccl} assumes a public dataset can be used to distill knowledge from global old models.

\smallskip \noindent \textbf{Replay-based Approaches.} The reason for the forgetting of learned tasks is that the model optimization is solely for new tasks, therefore, there is a question can we add objectives for learned tasks to the optimization? Accordingly, replay-based solutions are proposed, which replay old tasks during FCL. 
Intuitively, random storing can be inefficient, thus GLFC~\cite{glfc} tries to store the samples that are closest to each class center. In addition to such local storing of old class data, \cite{global_memory} builds a global memory bank hosted on the central server to better deal with the global forgetting caused by data heterogeneity. However, managing this global memory requires clients to upload their private data, which violates the privacy preservation of FL, thus \cite{global_memory} introduce Laplace noise to protect uploaded data, and other works leverage generative models like GAN~\cite{qi2022better} to generate old class data for replay. Besides, different from data sample replay, \cite{anonymous2023accurate} uses a normalization flow model to generate old task features for replay while learning new tasks. Similar ideas can also be found in~\cite{prototype_1}, which replay old class prototypes across clients and the server. 

\smallskip \noindent \textbf{Architecture-based Approaches.} To preserve the old task knowledge, there is another intuitive way -- disentangling different tasks and distributing them to separate model parameters. FedWeIT~\cite{fedweit} decomposes model parameters into task-general and task-specific during federated training. The local-specific parameters are preserved unchanged in local training and uploaded to a parameter bank after an operation of sparsification. Then the sparse task-specific parameters can be selectively constituted for different tasks. FedSeIT~\cite{fedseit}, a natural language processing study, also divides the model parameter into separate parts. Except for similar task-general and specific parts, there is another part to measure the correlation between tasks in FedSeIT, thus it can spare the parameter bank. Similar ideas~\cite{prompt_1} can be regarded as decomposing prompts for different tasks and updating the prompt space of the global model to achieve FCL. FOT~\cite{fot} modifies the new task training to make its model updates orthogonal to the previous task activation principal subspace, which can prevent interference between tasks. 

\subsection{Evaluation}
\noindent \textbf{Metrics.} Two metrics are used to measure the performance of FCL -- average accuracy (ACC) and forgetting (FGT)~\cite{fot}, 
\begin{align}
\mathrm{ACC} = \frac{1}{t}\sum_{i=1}^t a^{i, t},\quad \mathrm{FGT} = \frac{1}{t-1} \sum_{i=1}^{t-1} a^{i, i} - a^{i, t},
\end{align}
where $a^{i, t}$ is the model accuracy of task $\mathcal{Y}^i$ right after learning task $\mathcal{Y}^t$. As for Asynchronous FCL, ACC and FGT of local clients are also used to measure the performance.

\section{Future Directions}
\label{sec_future_direction}
\subsection{Future Opportunities of Incorporating New Features and Tasks into FL}
\label{sec_future_features_tasks}
\noindent \textbf{More Setups.} For incorporating new features, a feasible exploration will unify FDA and FDG to achieve continuously evolving FL. After adapting to more and more domains, the evolving FL model becomes more generalizable to any unseen domains. As for the incorporation of new tasks, future work needs to study the multi-label and multi-grained classification, task transition from classification to other tasks (semantic segmentation, object detection, etc.), and multi-modal learning for more complicated tasks (e.g., Visual Question Answering). Some dedicated setups for FCL are also worth studying, like how to design OOD detection for distinguishing new features belonging to seen or unseen tasks, and how to deal with FCL where both the task order and composition among clients are different.  

\smallskip \noindent \textbf{Efficiency.} Nearly all current studies of incorporating new knowledge into FL do not consider the efficiency concern, thus how to cut their computation overhead and communication cost needs further exploration. The domain-invariant learning can also be achieved by decomposing the features into domain-shared and domain-specific, and dedicated updating strategies are needed, which can help improve efficiency. For incorporating new tasks, splitting the model and training task modules with various strategies that depend on the task difficulty can also reduce the computation and communication overhead. As for FCL, prototype-based replay~\cite{prototype_1} has the potential to achieve more efficient learning, and communicating prototypes costs much less than communicating gradients. Note that it is also worth designing new incentive mechanisms with new metrics to evaluate the cost and gain of incorporating new knowledge, prompting proposals of more efficient techniques.

\subsection{Incorporate New Models}
\label{sec_new_models}
Integrating a broader range of advanced models, including Foundation Models (FM) and Large Language Models (LLM), into pre-existing FL systems, represents a significant enhancement over traditional FL frameworks~\cite{zhuang2023foundation}. As aforementioned, traditional FL systems often fall short in managing complex data distributions and meeting the demands of sophisticated tasks. The inclusion of models with superior architecture and larger scales is essential. This approach not only addresses the limitations of older FL systems but also enables evolution by increasing data availability, boosting collaborative development, and enhancing both the utility and privacy of FL models in various applications.
In the following context, we consider two cases for incorporating new models $\mathcal{M}_\mathrm{new}$ into FL: 1) leveraging new models (such as $f_\alpha$) to enhance existing models $f_\theta$; 2) conducting the architecture transfer (e.g., ResNet to Transformer) from the existing ($f_\theta$) to new models $f_\alpha$. In both these two cases, the new model $f_\alpha$ is assumed to be pre-trained on certain datasets. In the first case, clients may use their local data to distill the knowledge from $f_\alpha$ that is usually hosted at the central server $\mathcal{S}$ ($f_\alpha$ is usually quite large). As for the second case, the major effort lies in efficiently transferring old knowledge of $f_\theta$ to prepare a better optimization start for $f_\alpha$. 

\smallskip \noindent \textbf{Methods.}
We review the potential advances in the two according strategies. For the former, two major issues need to be considered: 1) using local client data is insufficient to take full use of the knowledge embedded in new models; 2) new models are usually too large to load on clients or communicate between clients and the server. To deal with these issues, some preliminary studies may be useful. For instance, CROSSLM~\cite{deng2023mutual} illustrates the use of small language models to refine LLMs and their generative capabilities to create synthetic data, facilitating data-free knowledge transfer that addresses resource and proprietary limitations in FL. Additionally, approaches such as using public data and LLMs for the differential private training of FL models through knowledge distillation, as investigated by~\cite{wang2023can,shao2024selective}, which enhances federated distillation by accurately selecting knowledge from local and ensemble predictions, show the potential for knowledge transfer from new models to established FL models. In future research, more efforts are needed to address the second issue, like communicating proxy small models among clients and the server or enabling secure inference of FM or LLM hosted on the server for local client data. In addition, it is also important to enhance the flexibility of incorporating new models, including the best time to incorporate what models, and whether the knowledge in new models needs to be decomposed for task-general or specific incorporation.

As for the latter case -- architecture transfers from the old to the new models, existing FL studies focusing on heterogeneous model architectures may help. For instance, FedRolex~\cite{alam2022fedrolex} enables a rolling window mechanism for clients to selectively extract sub-models from the global model hosted on the central server, and then the extracted sub-models are specific to the training of different local models. Similar ideas can be found in \cite{ye2023heterogeneous}. There is another potential work~\cite{fccl} that leverages a public dataset and SSL to distill knowledge from the old FL models to new models with heterogeneous architectures, but the public dataset is required to be similar to the learned data distribution, which is unreasonable in practice. In future studies, the data will always be a big issue, i.e., maybe not all learned data can be stored for use during the transfer to new architectures, in this case, data selective storing is needed. Besides, if the new architectures are relatively large, a dedicated design of splitting learning or tuning is needed, such as instructing adapters associated with different attention layers for training transformers.


\subsection{Incorporate New Aggregation Algorithms}
\label{sec_new_algorithms}
Existing FL aggregation algorithms can be categorized based on their focus: those that address data heterogeneity and those that prioritize data privacy protection. Representative algorithms of the former include FedProx~\cite{fedprox}, Scaffold~\cite{scaffold}, and Moon~\cite{moon}, which are designed to incorporate more considerations for data heterogeneity in local optimization. For the latter, the main methodology involves using secure Multi-Party Computation (MPC)~\cite{secure_mpc} to transform various representations of clients' data that need to be communicated — including model gradients, intermediate neural representations, and inference predictions — into secrets for distributed sharing and computation. Typical works in this area are SecAgg~\cite{secagg} and CaPC~\cite{capc}. In future research, one important exploration could be combining these two types of aggregation algorithms to effectively deal with data heterogeneity while ensuring strict cryptographic data privacy protection. Additionally, in our focus on FL with new knowledge, more urgent research should be on how to transition from old to new algorithms, especially assuming that the current FL model is trained with the old algorithm. In this direction, future research should address the following questions: 1) Whether directly switching aggregation algorithms could slow down the convergence speed of the FL model or impair its final convergence performance; 2) If there is an optimal timing for switching algorithms and how to determine this timing; 3) How to handle scenarios where the new algorithm requires communication of additional historical information, which was not stored during the training process of the old algorithm.

\subsection{New Threats of FL with New Knowledge}
\label{sec_new_threats}
In addition to model parameter updates or gradients, there are usually additional contents that need to be communicated when incorporating new knowledge. Therefore, it is urgent to propose dedicated methods to protect these additional contents in the semi-honest scenarios, in particular, against a variety of inference attacks~\cite{wang2019eavesdrop,inference_attack}. Some preliminary tries include adding random~\cite{glfc} or differential private noise~\cite{global_memory} to the domain or task-specific models or class prototypes of FCL. However, such privacy protection lacks theoretical security guarantees and hurts the performance of FL systems. Aligning with the intrinsic nature of multiple parties in FL, it may be possible to leverage MPC to communicate the additional contents. MPC is built on cryptography and there are strong security guarantees. Besides, because the carriers of new knowledge are features, tasks, models, and algorithms, we need to consider the worst case that these carriers are polluted or poisoned in malicious scenarios. For instance, features with poison attacks may be disguised as new features, and if there is no detection tool, they will be naturally incorporated in FL. Furthermore, malicious clients may conduct unintended local model training for society-harmful tasks, in this case, there is a need to distinguish them from new benign tasks. New models may also contain backdoors~\cite{xie2020dba,guo2023policycleanse}, blindly incorporating them will cause serious outcomes. New FL algorithms also expose new platforms for new attacks, like gradient inversion attacks~\cite{inversionattack} are only effective to certain FL algorithms.

\section{Conclusion}
In this paper, we focus on incorporating new knowledge into FL to achieve its sustainable development. We consider new knowledge from four sources: features, tasks, models, and algorithms. For incorporating new features, we discuss a set of processes to enhance the out-of-distribution generalization ability of FL systems. Regarding new tasks, we discuss how to use task-personalized and self-supervised FL to strengthen the cross-task generalization capability of FL, and we also introduce how to learn new tasks through federated continual learning. Additionally, we review existing research that may facilitate the fusion of different FL models and the transition between different algorithms. Finally, we comprehensively discuss future research directions, considering factors such as scenario setups, efficiency, and security.


\bibliographystyle{named}
\bibliography{ijcai24, ijcai24_djh}

\end{document}